\documentclass{article}
\usepackage{spconf,amsmath,graphicx}

\usepackage[utf8]{inputenc}
\usepackage{booktabs}
\usepackage{multicol}
\usepackage{graphicx}
\usepackage{multirow}
\usepackage{amssymb,amsmath,bm}
\usepackage{enumitem}
\usepackage{soul,color}
\usepackage{array}
\usepackage{footnote}
\makesavenoteenv{table}
\usepackage{hyperref}
\usepackage{color}
\usepackage{graphicx}
\usepackage{subfigure}

\usepackage[left]{lineno}
\usepackage[utf8]{inputenc}
\usepackage{amsmath,bm}
\usepackage[ruled, linesnumbered]{algorithm2e}
\newcommand{\bs}[1]{\boldsymbol{#1}}

\title{End-to-end contextual speech recognition using class language models and a token passing decoder}
\name{Zhehuai Chen$^{* 1,2}$, Mahaveer Jain$^2$, Yongqiang Wang$^2$, Michael L. Seltzer$^2$, Christian Fuegen$^2$
\thanks{
$^*$This work was done when the first author was an intern with Facebook.
}
}
\address{
  $^1$SpeechLab, Department of Computer Science and Engineering, Shanghai Jiao Tong University\\
  $^2$Facebook, One Hacker Way, Menlo Park, CA 94025, USA}
	\email{chenzhehuai@sjtu.edu.cn, \{jainmahaveer,yqw,mikeseltzer,fuegen\}@fb.com}

\begin{document}
\ninept
\maketitle
\begin{abstract}

End-to-end modeling (E2E) of automatic speech recognition (ASR) blends all the components of a traditional speech recognition system into a unified model.
Although it simplifies training and decoding pipelines,
the unified model is hard to adapt when mismatch exists between training and test data. In this work, we focus on contextual speech recognition, which is particularly challenging for E2E models because it introduces significant mismatch between training and test data. To improve the performance in the presence of complex contextual information, we propose to use class-based language models(CLM) that can populate the classes with context-dependent information in real-time. To enable this approach to scale to a large number of class members and  minimize search errors, we propose a token passing decoder with efficient token recombination for E2E systems for the first time.
We evaluate the proposed system on general and contextual ASR, and achieve relative 62\% Word Error Rate(WER) reduction for contextual ASR without hurting performance for general ASR.
We show that the proposed method performs well without modification of the decoding hyper-parameters across tasks, making it a general solution for E2E ASR.

\end{abstract}
\begin{keywords}
End-to-end Speech Recognition, Weighted Finite State Transducer, Token Passing, Class-based Language Model
\end{keywords}

\vspace{-0.6em}
\section{Introduction}
\label{sec:intro}
\vspace{-0.6em}
Automatic speech recognition (ASR) with Deep Neural Networks (DNN) commonly operates
in the hybrid framework: DNNs, as discriminative acoustic models (AM),
estimate the posterior probabilities of Hidden Markov Model (HMM)
states.  
In inference stage, external lexicons and language models (LM) are combined with acoustic models (AM). All of these models are optimized independently~\cite{hinton2012deep}.
The {\em Weighted Finite State Transducer} (WFST) paradigm ~\cite{mohri2002weighted} has been proposed
to combine different knowledge sources and perform search space optimization to achieve efficient decoding, that is, to find the the sequence of labels that best match the input audio with minimum search errors.
%
One way to perform efficient search is via {\em Token Passing}, which is a single-pass algorithm that can generate multiple alternatives for each WFST state~\cite{young1989token}.

Recently proposed E2E speech recognition has arisen as a result of both recent advances in neural modeling of context and history in sequences~\cite{sak2014long,chen2018progressive}, and access to more labeled training data for better generalization. In E2E speech recognition, a single model predicts words directly from acoustics, unifying the acoustic, language, and pronunciation models into one system. 
Although E2E training benefits from sequence modeling and simplified inference~\cite{zhc00-chen-tasl2017,chen2017unified}, 
it performs worse than traditional systems in long and noisy speech recognition and needs a larger amount of transcribed acoustic data~\cite{soltau2016neural,huang2018ctc} to perform well. Moreover, traditional language models  and the corresponding decoding techniques are difficult to incorporate into E2E systems~\cite{hori2017multi,zhc00-chen-icassp17-e2e}.
The inability to exploit knowledge from external language models and lexicons especially hampers the adaptability of E2E systems. 

This is particularly important for {\em Contextual Speech Recognition}~\cite{mcgraw2016personalized} where user context can provide additional information on likely query clues as to what the user will say.
%
%
The prior work in this field suffers from limited extendibility in both context complexity and the amount of context phrases~\cite{williams2018contextual,pundak2018deep}, discussed in Section~\ref{sec:review_ctx_e2e}.

In this work we propose to model the context-specific information using a class-based LM (CLM)~\cite{kneser1993improved}. In this paradigm, contextual knowledge is modeled by an $n$-gram LM. Context phrases are composed on-the-fly into the CLM based WFST~\cite{hall2015composition}. 
Because the WFST is non-deterministic, as  discussed in Section~\ref{sec:tok_pass}, each E2E inference hypothesis includes alternative paths in the WFST. To handle these alternative paths, we propose a token passing decoder with efficient token recombination for E2E systems for the first time.

The rest of the paper is organized as follows. 
In Section~\ref{sec:review_ctx_e2e}, prior works in E2E contextual speech recognition are briefly reviewed. Our main contributions are
in Sections~\ref{sec:proposed}: i) use CLM to solve contextual speech recognition and wake word  problem. ii) propose a token passing decoder for E2E inference. 
The relation to prior work is discussed in Section~\ref{sec:prior}.
Experimental results are presented in Section~\ref{sec:exp}, followed by conclusion in Section~\ref{sec:conclu}.

\vspace{-0.6em}
\section{Contextual E2E Speech Recognition}
\label{sec:review_ctx_e2e}
\vspace{-0.6em}
\subsection{Attention-based End-to-end Modeling}
\label{sec:review_e2e}
\vspace{-0.6em}
We use the attention-based encoder-decoder model~\cite{chan2016end} for E2E modeling.
It predicts the posterior probability of label sequences given both a feature sequence $\mathbf{x}$ and previous inference labels $\mathbf{l}_{1:i-1}$.
\begin{equation}
\vspace{-0.6em}  
\label{equ:enc-dec}
\begin{split}
P(\mathbf{l}|\mathbf{x})=\prod_i P(l_{i}|\mathbf{x},\mathbf{l}_{1:i-1})
\end{split}
\end{equation}
\begin{equation}
\label{equ:enc-dec-hid}
\begin{split}
\mathbf{h}&=\text{Encoder}(\mathbf{x})\\
\end{split}
\vspace{-0.3em}  
\end{equation}
\begin{equation}
\label{equ:enc-dec-dec}
\vspace{-0.1em}  
\begin{split}
P(l_{i}|\mathbf{x},\mathbf{l}_{1:i-1})
&=\text{AttentionDecoder}(\mathbf{h},\bs s_{i-1})
\end{split}
\end{equation}
where $\mathbf{h}$ is the sequence of encoder states and $\mathbf{s_{i-1}}$ is the decoder hidden state from the previous time step. The $\text{Encoder}(\cdot)$ is typically a unidirectional or bidirectionaly long short term memory  (LSTM) network while the $\text{AttentionDecoder}(\cdot)$ is a unidirectional LSTM. 

Compared to traditional decoder in a hybrid system~\cite{hinton2012deep}, the $\text{AttentionDecoder}(\cdot)$ implicitly captures LM information in a way that is jointly trained with the $\text{Encoder}$ which can be interpreted as an acoustic model. 
%
%
%
Because of this tight unification between models and decoder, such E2E systems cannot be easily adapted to new domains or contexts. In contrast, traditional systems can do this easily via updates to the language model~\cite{mcgraw2016personalized}.  
\vspace{-0.6em}
\subsection{On-the-fly Rescoring with External WFST}
\label{sec:review_otf}
\vspace{-0.6em}
A contextual automatic speech recognition (ASR) system dynamically
incorporates real-time  context into the recognition process of a speech recognition system~\cite{mcgraw2016personalized}. A typical example of contextual information is the 
personal information such as a user's contacts.

One branch of methods is to generate an on-the-fly contextual LM and include it into the recognition process to bias the beam search in Section~\ref{sec:review_e2e}.
\cite{williams2018contextual} introduces the shallow fusion approach.
\begin{equation}
\vspace{-0.6em}
\label{equ:shallow}
\begin{split}
\mathbf{l}^*= \mathop{\arg\max}_{\mathbf{l}}\log P(\mathbf{l}|\mathbf{x}) + \lambda\log P_C(\mathbf{l})
\end{split}
\end{equation}
where $P_C(\mathbf{l})$ is the introduced contextual LM and $\lambda$ is a scaling factor.
\cite{williams2018contextual} proposes to use similar on-the-fly rescoring technique as~\cite{hall2015composition} to obtain  $P_C(\mathbf{l})$. The method shows good performance in limited number of context phrases but the recognition accuracy starts to drop when the number of contexual phrases is above 100.
One possible reason could be that it follows the WFST search idea from~\cite{hall2015composition} to traverse epsilon arcs only in the absence of a matching symbol. This can introduce significant search errors inside the word class~\footnote{\cite{hall2015composition} introduces a special backoff method on  word level. Beside that, \cite{williams2018contextual} does not describe any further design on its "speller" WFST inside the word.}.  %
We will look into this problem and  propose  methods to alleviate it  in Section~\ref{sec:tok_pass}.  
\vspace{-0.6em}
\subsection{Contextual E2E Modeling}
\label{sec:review_deep_ctx}
\vspace{-0.6em}
Another method that try to integrate contextual information into the E2E modeling is called CLAS~\cite{pundak2018deep}. This technique  first embeds each 
phrase, represented as a sequence of graphemes, into a fixed-dimensional representation. And then it employs an attention
mechanism to summarize the available context at each step of the output predictions.
By this way, CLAS explicitly models the probability of seeing particular phrases given audio and
previous labels.

To scale up this paradigm and make it into use, two fundamental problems need to be considered: i) Model the similarities between large amounts of context phrases. Although~\cite{pundak2018deep} proposes a conditioning mechanism to reduce the amount of phrases considered, a better and more unified solution is important to the scaling up~\cite{pundak2018deep}.
ii) Constrain the search space of context phrases  at a particular step in $\text{AttentionDecoder}(\cdot)$. The above attention mechanism is done by using all phrases, while general CLM~\cite{kneser1993improved} applied in this paper, only uses contextual phrases that are relevant at current prediction step.

\vspace{-0.6em}
\section{The Proposed Method}
\label{sec:proposed}
\vspace{-0.6em}
\subsection{Class-based Language Model and WFST}
\label{sec:clm-wfst}
\vspace{-0.6em}

This work follows the paradigm and formulation of the shallow fusion in Section~\ref{sec:review_otf}. 
To solve the extendibility in modeling complex context, we first extend the paradigm by using CLM~\cite{kneser1993improved}.
CLM refers to introducing word equivalence classes into  $n$-gram LM. 
In contextual speech recognition, the contextual phrases, e.g. a user’s favorite
songs and contacts, can be grouped into multiple word equivalence classes ({\em inside the class}). And the context of the conversation is modeled by $n$-gram LM ({\em outside the class}). 

\begin{figure}
  \centering
  \vspace{-0.6em}
    \includegraphics[width=0.8\linewidth]{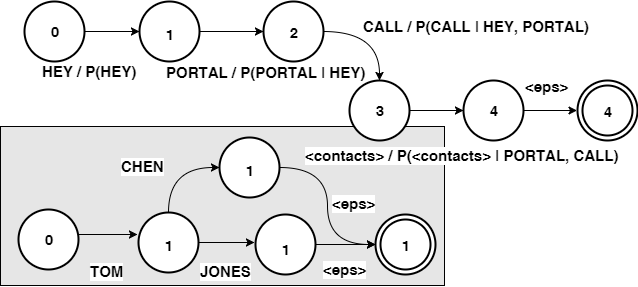}
    \caption{\it Examples of CLM in Contextual Speech Recognition (word level WFSTs for simplicity). "HEY / P(HEY)" denotes the symbol of arc is "HEY" with the probability P(HEY).  States in the shaded box represent names in the word class "$\langle contact \rangle$". }
    \label{fig:exp-classlm}
    \vspace{-1.5em}
\end{figure}

We compile $n$-gram contexts with word equivalence classes(call @name) and, contextual phrases(Tom Cruise, Lady Gaga) of each word equivalence class in separate WFST graphs. These WFST graphs are then composed with the "speller" WFSTs~\cite{williams2018contextual} to obtain the grapheme level WFSTs. We do determinization operation on WFSTs of contextual phrases to reduce number of tokens as discussed in the next section.
In the inference stage, a form of on-the-fly composition between the {\em inside} and {\em outside the class} WFSTs is conducted~\cite{hall2015composition}, without requiring any changes to the pre-compiled transducers~\cite{dixon2012specialized}. Figure~\ref{fig:exp-classlm} shows an example.

%
%
%
%

Purposed framework can also improve the wake word recognition in the E2E system. We add a special word class in the start of the sentence with a boosting factor ({\em keyword boosting}; tuned on the development set). In the inference stage, the  wake word grapheme sequence is composed into word equivalence class similar to context phrases.


\vspace{-0.6em}
\subsection{Token Passing Decoder}
\label{sec:tok_pass}
\vspace{-0.6em}

In~\cite{hall2015composition}, they assume that weight of matched-symbol arcs would always be lower than backoff arcs. Because of this assumption, $n$-gram LM WFST can be treated as deterministic and a single token decoder can be used. 

\begin{figure}[ht]
  \centering
  \vspace{-1.5em}
    \includegraphics[width=\linewidth]{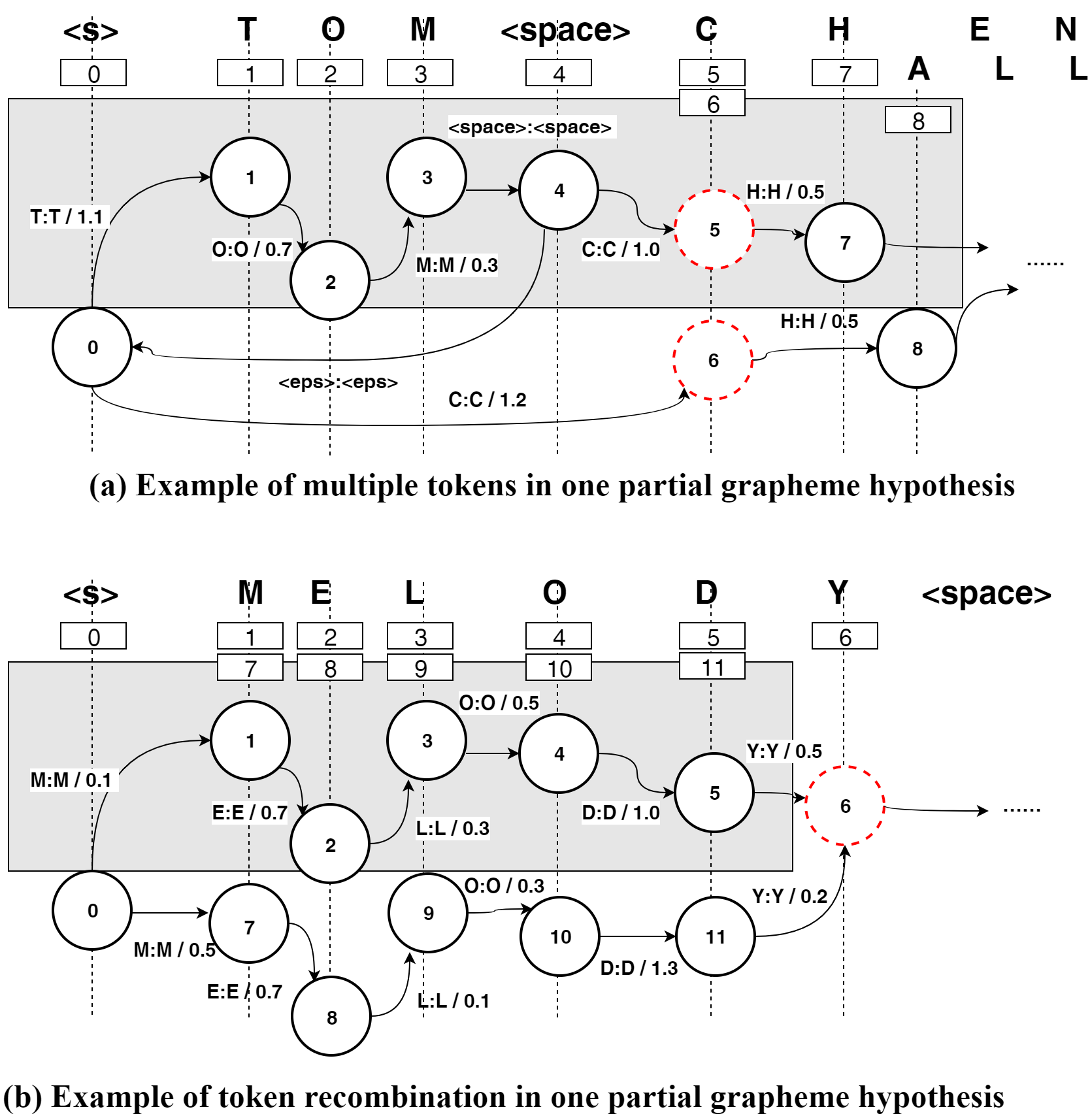}
    \caption{\it Examples of the Token Passing Algorithm. States in the shaded box are of the word class and they are generated from on-the-fly composition. The small boxes below graphemes denote their tokens and state numbers. In (a), 
   the sixth grapheme "C" is the prefix of "CHEN" and "CALL", corresponded to two tokens in different WFST states (red dash circles). 
   In (b), there are two "MELODY" in both {\em inside the class} and {\em outside the class}. Their tokens can be recombined in state 6, where they are with the same history states in both E2E and WFST.  }
    \label{fig:exp-tok-pass}
    \vspace{-1.5em}
\end{figure}

Grapheme level WFST is non-deterministic because of following reasons:
i)  $n$-gram LM has backoff transitions.
ii) Duplication of phrases between two word equivalnce classes.
iii) Duplication of phrases between a word equivalnce class and $n$-gram LM words. ii) and iii) are issues only because of  on-the-fly composition as we never determinize the whole WFST. Figure~\ref{fig:exp-tok-pass}(a) shows an example. Because of non-deterministic WFST, each graphemes hypothesis of E2E inference can have multiple paths in WFST. %
To cope with multiple tokens for a hypothesis, we purpose token passing decoder for E2E system in Algorithm~\ref{code:tokpass}, with efficient token recombination. Figure~\ref{fig:exp-tok-pass} shows examples of how to process tokens in our algorithm.

In this algorithm, the $k$-th token in the token set $\mathcal{H}$ is defined as a 4-element tuple $(\bs s_k, \mathbf{l}_k, t_k, q_k)$, whereas $\bs s_k$ is the $k$-th decoder hidden state $\bs s_{i-1}^{(k)}$ at the last prediction step in Equation~(\ref{equ:enc-dec-dec}); $\mathbf{l}_k$ is the partially-decoded sequences; $t_k$ is the last state of WFST path whose output sequence is $\mathbf{l}_k$. Note that as discussed above, there could be multiple $t_k$ associated with the same $\mathbf{l}$; $q_k$ is the score for the $k$-th partial hypothesis. At each decoding step, we expand $k$-th token by concatenating $\mathbf{l}_k$ with every grapheme: $\bs s_k$ and $q_k$ are first updated in Line 9 and 10. WFST states are then expanded in Line 12 to 15: $\mathrm{SearchFST}(t_k, l)$ returns all the possible WFST states (and the correspoding cost) which can be reached by departing from the state $t_k$ and consuming the input symbol $l$; again multiple tokens can exist with the new hypothesis output $\mathbf{l}'$. 
$\mathrm{TokenRecombination}$ function is proposed in Line 16:
for every $(\mathbf{l}_k', t_k')$ pair, it is only necessary to maintain the best token;  for the same partial hypothesis $\mathbf{l}_k'$, we only maintain $B_{\tt tok}$ tokens at most. 
Finally, after expanding all the current tokens in $\mathcal{H}$, we select the best $B$ partial hypotheses $\mathbf{l}_k'$ in SelectTopN function. For the same partial hypothesis, we need to maintain multiple tokens with different WFST states in our purposed algorithm but its time complexity  is similar to standard beam search as discussed later.  Compared to the on-the-fly rescoring~\cite{williams2018contextual}, key differences of our algorithm are:
\begin{itemize}[leftmargin=*]
\item Multiple tokens. 
In the WFST search (line 12), 
previously purposed on-the-fly rescoring ~\cite{hall2015composition} traverses epsilon transitions only in the absence of a matching symbol. Figure~\ref{fig:exp-tok-pass}(a) shows an example where this can introduce search errors. For the grapheme "C",  our proposed method would have two tokens corresponding to state 5 and 6. State 6 extends from the backoff state 0, which can be traversed  from state 4. In~\cite{williams2018contextual}, state 0 is not traversed because state 4 already has a matched arc to state 5. This results in not exploring state 6 and hence introducing a possible search error.
In this work, we propose to keep multiple tokens for each hypothesis from E2E inference using token passing method.

\begin{algorithm}
\label{code:tokpass}
\textbf{Input:} $\mathbf{h}$, defined in Equation~(\ref{equ:enc-dec-hid})  \\
\textbf{Initialization:} 
    $\mathcal{H}=\{ (\bs s_0, \texttt{"<bos>"}, \texttt{FST.start}, 0)\}$ ;

    \While{$\mathrm{EndDetection}(\mathcal{H})$~\cite{watanabe2017hybrid}~\protect\footnotemark} {
        $\mathcal{H}' \leftarrow \{\}$\;
        \For{$(\bs s_k, \mathbf{l}_k, t_k, q_k) \in \mathcal{H}$} {
            \For{ each grapheme $l$}{
                $\mathcal{H}_{l} = \{\}$ \;
                $\bullet$ \textrm{extend decoder network by grapheme l}\\
                \Indp
                    $\mathbf{l}_k' \leftarrow \mathbf{l}_k + l$;  $ q_k' \leftarrow q_k + p(l | \bs s_k, \mathbf{h})$\;
                    $\bs s_k' \leftarrow \textrm{UpdateDecoderState}(\bs s_k, l)$ \;
                \Indm
                $\bullet$ \textrm{extend FST state}\\
                \Indp
                	$\mathcal{T}_k' \leftarrow \mathrm{SearchFST}(t_k, l)$\;
                    \For {$(t_k', p_k') \in \mathcal{T}_k'$} {
                        $\mathcal{H}_l \leftarrow  \mathcal{H}_l \cup \{(\bs s_k', \mathbf{l}_k', t_k', q_k'+ \lambda p_k') \}$\;
                    }
                \Indm
                $\mathcal{H}_l \leftarrow \mathrm{TokenRecombination}(\mathcal{H}_l, B_{\texttt{tok}})$\;
                $\mathcal{H}' \leftarrow \mathcal{H}' \cup \mathcal{H}_l$
            }
        }
        $\mathcal{H} \leftarrow  \mathrm{SelectTopN}(\mathcal{H}', B)$
    }
    return best path in $\mathcal{H}$\;
\caption{Token Passing Algorithm for E2E Model}
\end{algorithm}
\footnotetext{We also force hypotheses to end in the end of WFST.}

\item Token Recombination. To reduce the amount of tokens in each hypothesis  and add diversity in WFST paths, token recombination is proposed for the decoder. Tokens can be combined only if they have both the same WFST state and  E2E state. As different hypotheses have different E2E states, the token recombination can only be conducted on tokens of one hypothesis, e.g. Figure~\ref{fig:exp-tok-pass}(b).

\end{itemize}

The proposed method does not change the computational complexity of standard beam search in E2E inference. The complexity of standard E2E inference is $C_{E2E}=O(B\cdot L\cdot D) + O(F) $, where $B$ is the beam size of hypotheses , $L$ is the length of the sequence, $D$ is the complexity of the decoder neural network of E2E and, F is the complexity of encoder neural network of E2E. The token passing decoding does not change the beam search in E2E inference. Complexity of token passing decoding is $C_{DEC}=O(B\cdot L \cdot B_{tok} \cdot U)$,
where $B_{tok}$ is the beam of WFST tokens in each hypothesis,  $U$ is the size of grapheme. Since $D \gg B_{tok} \cdot U$, we have $C_{DEC} + C_{E2E} \approx C_{E2E}$. 

\vspace{-0.6em}
\section{RELATION TO PRIOR WORK}
\label{sec:prior}
\vspace{-0.6em}

In the inference stage of the E2E speech recognition, prior work such as ~\cite{zhc00-chen-is16,chen2018search,chen2018gpu,kannan2017analysis,hori2017multi,wu2017multi} uses $n$-gram LM or NNLM to bias search space. 
In contextual E2E speech recognition, because of the mismatch between training and test utterances, it is even more important to integrate external knowledge sources to improve the WER.
\cite{williams2018contextual} proposes to use on-the-fly composed external contextual LM to bias the
beam search of E2E inference.
Another branch of methods~\cite{pundak2018deep} tries to model the probability of seeing particular context phrase given audio and previous labels.
The prior work in this field suffer from limited extendibility in both
context complexity and the amount of context phrases.
The advantages of this work include: i) Better generalization of complex context by using CLM. ii) Less search errors by keeping E2E and WFST states separately using multiple tokens with recombination in 1-pass decoding. 


\vspace{-0.6em}
\section{Experiments}
\label{sec:exp}
\vspace{-0.6em}
\vspace{-0.6em}
\subsection{Setup}
\label{sec:exp-setup}
\vspace{-0.6em}

The data is collected with the help of crowd sourced workers. These workers were asked to write and speak utterances that they could ask an AI assistant. Each utterance could belong to general speech~\footnote{e.g. what are the ingredient in pork stew?} or one of many possible domains~\footnote{e.g. play Lady Gaga song (music domain), call Alex (calling domain).}.  We have 10 million utterances for training. Size of {\em General} ASR testset is 60 thousand whereas size of {\em Contextual} ASR testset is 10 thousand. 
%
In order to generate possible members of CLM for {\em Contextual} ASR testset, we first extract the true entity from the utterance and then add 999 fake entities of the same type. Each utterance of {\em Contextual} ASR testset has wakeup word in the beginning.  

In training,
40-dimensional fiterbank feature is extracted with a frame rate of 10ms. E2E models with grapheme units were built with PyTorch~\cite{paszke2017pytorch} based on Espnet~\cite{watanabe2018espnet}. 2-layer BLSTM with 1400 nodes is used for the encoder, and 2-layer LSTM with 700 nodes is used for the decoder. The model is optimized by both connectionist temporal classification (CTC) and E2E criteria~\cite{watanabe2017hybrid}. 
 Our $n$-gram LM is 3-gram LM trained on vocabulary of 300 thousand words. 
We use value of 10 for both $B$ (hypothesis beam size) and $B_{tok}$ (WFST token beam size). 
Word error rate (WER) is used as the metric for evaluation.

\vspace{-0.6em}
\subsection{Performance Comparison}
\label{sec:exp-perf}
\vspace{-0.6em}

We compare performance of the E2E system for both {\em Contextual} and {\em General} ASR testsets. 
Decoding hyper-parameters used for both testsets are same in each of the row in Table~\ref{tab:perf-compare}.
First row shows experiments without using any LM for E2E system. The WER for {\em General} ASR testset is 5.9 whereas WER for {\em Contextual}  ASR test is 34. The hardness of contextual speech recognition stems from: i) Lack of wake word modeling in training set. This affects both the recognition of wake word and history modeling~\footnote{If the model does not see the wake word during training, it cannot recognize both the wake word and speech after the wake word while decoding because of the unseen history state} for the remaining words. ii) Lack of contextual phrases in training utterances.

\begin{table}[thbp!]
\vspace{-1.5em}
  \caption{\label{tab:perf-compare} {\it  WER of End-to-end ASR in General and Contextual ASR.  } }
  \vspace{0.21em}
  \centerline{
    \begin{tabular}{m{7em}  || c  c }
      \toprule
      system  &  General  & Contextual  \\
      \midrule
		\ E2E & 5.9 &  35.1 \\
        \midrule
		\ \ \ + $n$-gram LM& {\bf5.6} &  31.4 \\
		\ \ \ + Class LM&  5.7 & {\bf 13.5} \\
	\bottomrule
    \end{tabular}
  }
  \vspace{-0.6em}
\end{table}

The second row shows experiments with an external 3-gram LM for E2E system. Decoding is performed by the proposed token passing decoder. For {\em General} ASR testset,  WER improves to 5.6\% from 5.9\%. For {\em Contextual} ASR testset,  WER improves to 31.4\% from 35.1\% . The improvement of general ASR from external $n$-gram LM is consistent with the results in~\cite{kannan2017analysis}.  We do not examine NNLM as main purpose of this work is to improve the contextual ASR. Improvement for {\em Contextual} ASR testset results from boosting wake word as discussed in the end of Section~\ref{sec:clm-wfst}. Nevertheless, simply boosting the scores of wake word can only help the recognition of it, it does not solve the problem of history state mismatch of the remaining words. Traditional LSTM-HMM sytetm trained with cross-entropy criterion gets 5.6\% WER on  {\em General} ASR testset for same $n$-gram LM.

$n$-gram LM can easily be integrated with the CLM based paradigm as discussed in Section~\ref{sec:clm-wfst}. Experiemtns for the proposed CLM based token passing decoder is in the third row. It achieves similar performance~\footnote{The slight difference stems from more WFST branchings of contextual phrases in CLM based WFST.} as $n$-gram LM for {\em General} ASR testset but achieves significant improvement for {\em Contextual} ASR testset(from 31.4\% to 13.5\% WER). These improvements comes from 1) modelling context using CLM and 2) reducing search errors by token pass decoder.
We conduct more analysis in the next section.
CLM is also good at adaptability as shown by comparable results with $n$-gram LM for {\em General} ASR testset.

\vspace{-0.6em}
\subsection{Analysis}
\label{sec:exp-ana}
\vspace{-0.6em}

\begin{figure}[]
\vspace{-4em}
  \centering
    \includegraphics[width=\linewidth]{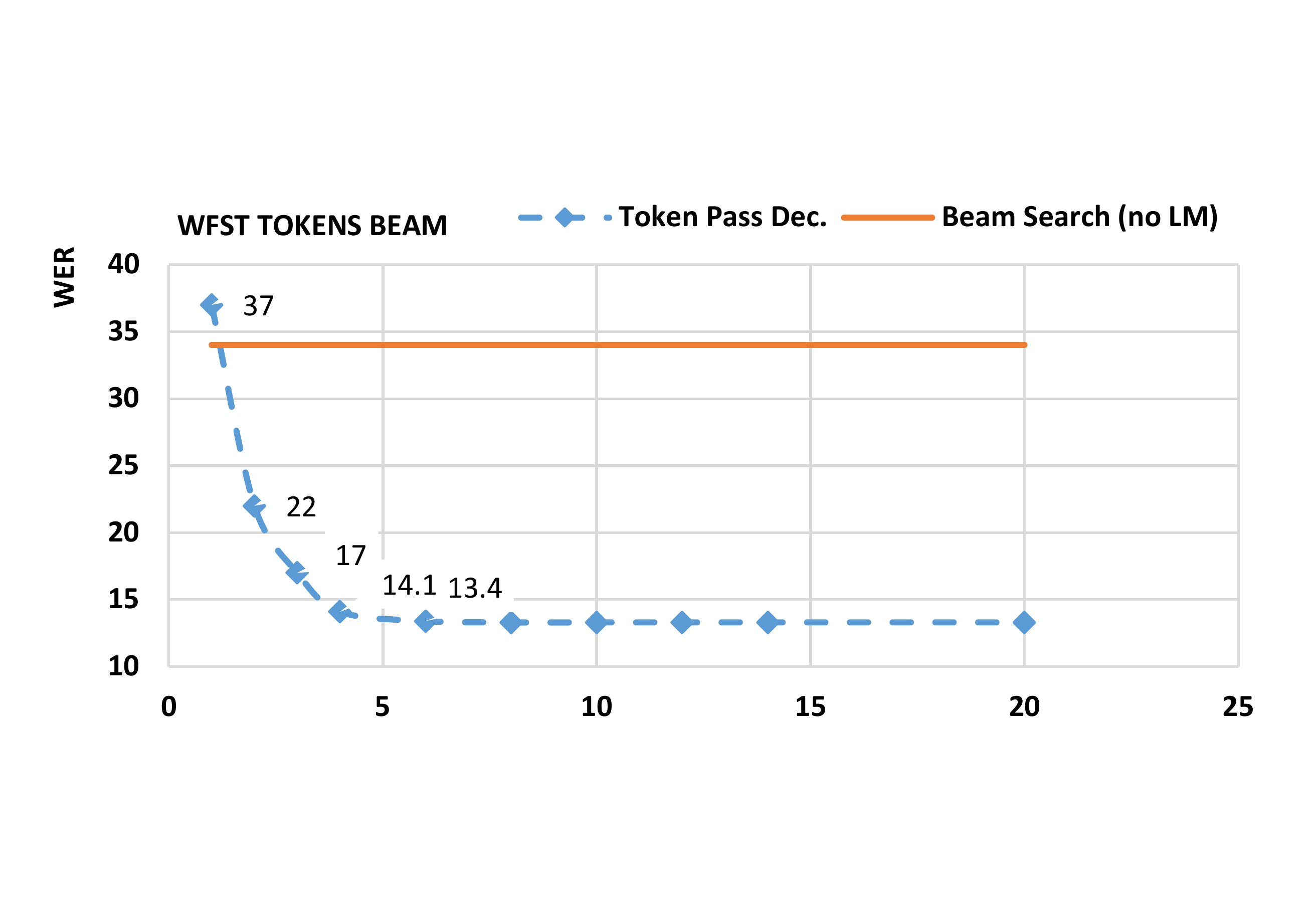}
    \vspace{-5.5em}
    \caption{\it WER v.s. Beam of WFST Tokens in the Token Passing Decoder.}
    \label{fig:exp-num-toks}
    \vspace{-1.5em}
\end{figure}

We firstly show the effectiveness of the proposed method compared to the shallow fusion ~\cite{williams2018contextual} based systems. Previous shallow fusion based systems essentially has value of $B_{tok}$(beam size of WFST tokens) as 1.
Figure~\ref{fig:exp-num-toks} shows relationship between WER and $B_{tok}$. With  $B_{tok}$ less than 5, the system has significantly worse performance~\footnote{Notably, we do not observe this phenomenon in general ASR~\cite{williams2018contextual}. We believe the different observation can be the over-biasing stems from the perplexity difference between word classes in CLM. }, which is consistent with the performance degradation with more than 1K phrases in~\cite{pundak2018deep}. This shows the importance of using multiple tokens in the WFST beam. 

\begin{figure}[ht]
\vspace{-4.5em}
  \centering
    \includegraphics[width=\linewidth]{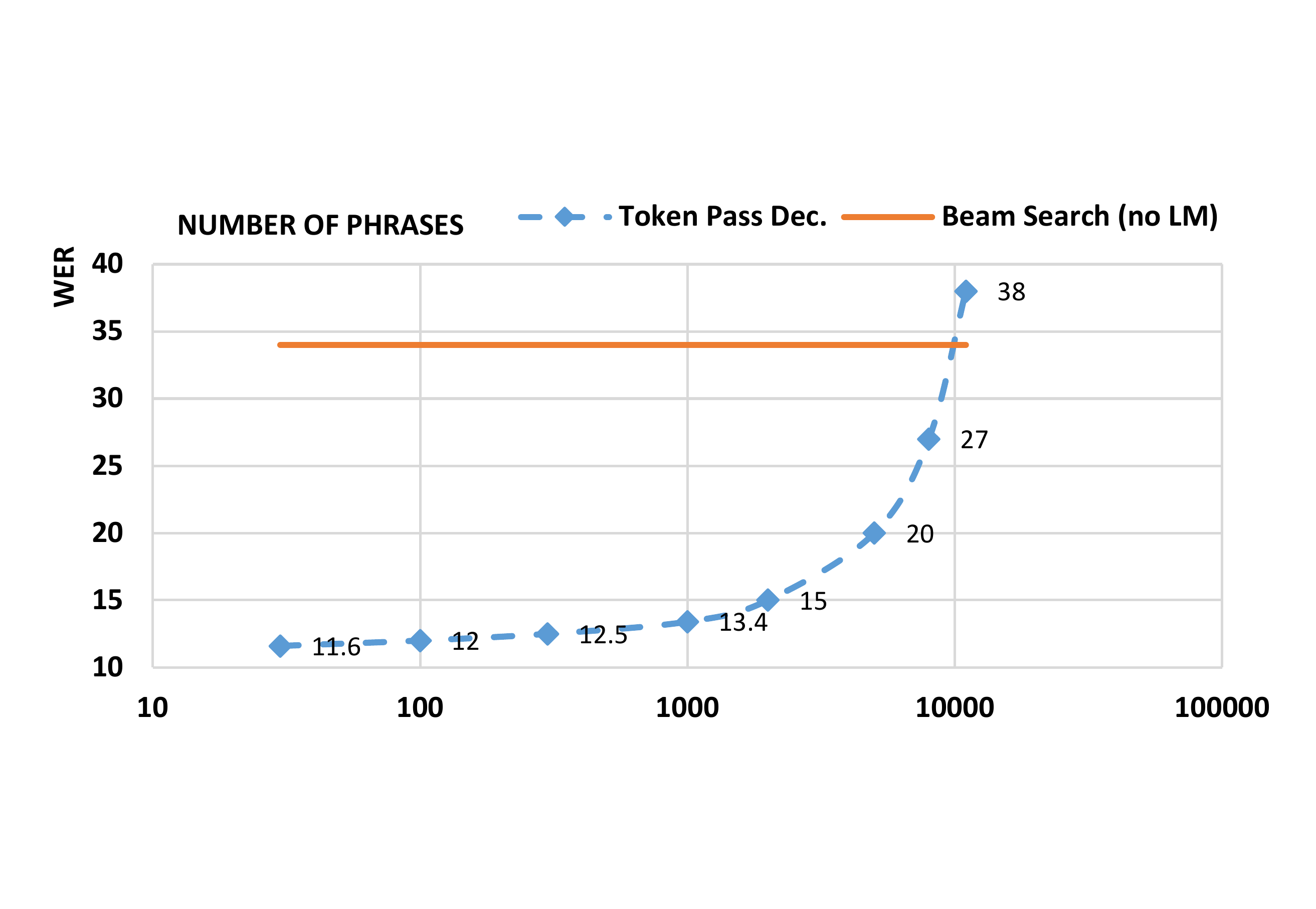}
    \vspace{-5.5em}
    \caption{\it WER v.s. Number of Phrases (beam of WFST tokens is 10). }
    \label{fig:exp-num-phrases}
    \vspace{-0.6em}
\end{figure}

In Figure~\ref{fig:exp-num-phrases}, we show experiments for scaling up number of contextual phrases. Though increasing the number of contextual phrases degrades the WER, we still have acceptable WER for upto 5K phrases, which is acceptable for most of the real world applications. After around 8K phrases, the system breaks down.

\begin{figure}[ht]
\vspace{-4.5em}
  \centering
    \includegraphics[width=\linewidth]{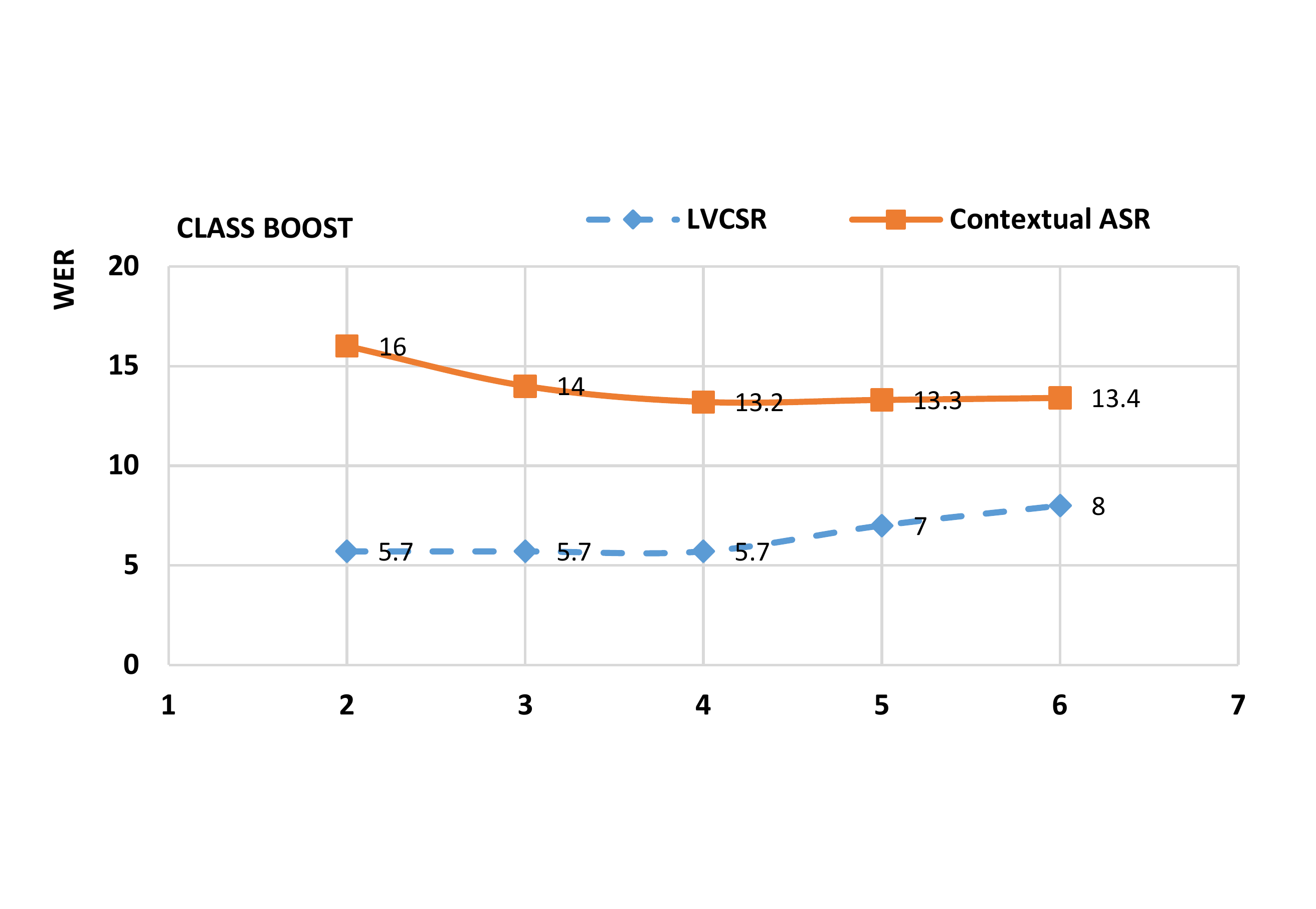}
    \vspace{-5.5em}
    \caption{{\em class boosting}, defined in Section~\ref{sec:clm-wfst} and its Effects in General and Contextual ASR. Similar trend in {\em keyword boosting}.}
    \label{fig:exp-ana}
    \vspace{-1em}
\end{figure}

Finally, we show impact on WER for {\em General } ASR when we tune hyper-parameter to improve {\em Contextual } ASR. The curves in Figure~\ref{fig:exp-ana} are mostly smooth, which shows general ASR performance is not sensitive to the contextual ASR performance and, vice versa.

\vspace{-0.6em}
\section{Conclusion}
\label{sec:conclu}
\vspace{-0.6em}

In this work, we propose to (a) use CLM to solve contextual speech recognition with (b) token passing decoder for E2E inference. 
The result on contextual ASR achieves consistent and significant improvements. Future works include extendibility to large number of context phrases and combining NNLM~\cite{li2018recurrent,zheng2016directed}.


\bibliographystyle{IEEEbib}
\bibliography{strings,refs}

\end{document}